\pdfoutput=1
\documentclass[journal]{IEEEtran}

\usepackage{times}
\usepackage{epsfig}
\usepackage{graphicx}
\usepackage{amsmath}
\usepackage{amssymb}
\usepackage{bbm}

\usepackage{cite}
\usepackage{float}
\usepackage{algorithm}
\usepackage{algorithmicx}
\usepackage{algpseudocode}
\usepackage{mathtools}
\usepackage{url}

\usepackage[table]{xcolor} 

\usepackage{threeparttable,booktabs}
\usepackage{etoolbox}
\appto\TPTnoteSettings{\footnotesize}

\usepackage{arydshln}
\usepackage{subcaption}
\usepackage{makecell}
\usepackage{pifont}
\usepackage{multirow}
\usepackage{booktabs}
\usepackage{chngcntr}
\usepackage{float}
\usepackage{diagbox}

\usepackage{amsmath}
\usepackage{romannum}

\makeatletter
\newcommand\threepart@subtable{
  \caption@setoptions{threepartsubtable}%
  \caption@ORI@threeparttable
}
\newenvironment{threepartsubtable}{%
  \threepart@subtable
}{%
  \endthreeparttable
}
\makeatother

\newcommand{\circled}[1]{\textcircled{\raisebox{-0.9pt}{#1}}}

\DeclareMathOperator*{\argmax}{arg\,max}

\makeatletter
\let\NAT@parse\undefined
\makeatother
\usepackage[breaklinks=true,bookmarks=false]{hyperref}
\hypersetup{
    colorlinks=true,
    linkcolor=blue,
    filecolor=magenta,
    urlcolor=cyan,
}

\title{\LARGE \bf
Gaze Preserving CycleGANs for Eyeglass Removal\\ \& Persistent Gaze Estimation 
}

\author{Akshay Rangesh$^\dag$, Bowen Zhang$^\dag$ and Mohan M. Trivedi\\
Laboratory for Intelligent \& Safe Automobiles, UC San Diego\\
{\tt\small \{arangesh, boz004, mtrivedi\}@ucsd.edu}
\thanks{
$^\dag$authors contributed equally
\newline Code, datasets \& models: \href{https://github.com/arangesh/GPCycleGAN}{https://github.com/arangesh/GPCycleGAN}
\newline \href{https://youtu.be/3_8U2TrrZVs}{Video results}
}
}

\begin{document}

\maketitle

\begin{abstract}
A driver's gaze is critical for determining their attention, state, situational awareness, and readiness to take over control from partially automated vehicles. Estimating the gaze direction is the most obvious way to gauge a driver's state under ideal conditions when limited to using non-intrusive imaging sensors. Unfortunately, the vehicular environment introduces a variety of challenges that are usually unaccounted for - harsh illumination, nighttime conditions, and reflective eyeglasses. Relying on head pose alone under such conditions can prove to be unreliable and erroneous. In this study, we offer solutions to address these problems encountered in the real world. To solve issues with lighting, we demonstrate that using an infrared camera with suitable equalization and normalization suffices. To handle eyeglasses and their corresponding artifacts, we adopt image-to-image translation using generative adversarial networks to pre-process images prior to gaze estimation. Our proposed Gaze Preserving CycleGAN (GPCycleGAN) is trained to preserve the driver's gaze while removing potential eyeglasses from face images. GPCycleGAN is based on the well-known CycleGAN approach - with the addition of a gaze classifier and a gaze consistency loss for additional supervision. Our approach exhibits improved performance, interpretability, robustness and superior qualitative results on challenging real-world datasets.
\end{abstract}


\section{Introduction}
Driver safety and accident risk are highly determined by a driver's attention levels, especially visual attention, e.g. a driver’s gaze, for both conventional and self-driving vehicles~\cite{distraction_Beanland_2013, Rangesh2018Awareness, Tawari2014Looking, trivedi2019attention, daily2017self}. It is crucial to have the ability to monitor the driver's gaze in real-time and use gaze information to efficiently enhance vehicle safety for intelligent vehicles. Previous work has proven that visual information from imaging sensors and corresponding learning algorithms are effective in determining driver gaze zones~\cite{Jha_2018, Martin_2018, Vora_2018, Yoon_2NIR_2019} and driver activity in general~\cite{deo2019looking, roitberg2020open, drive_and_act_2019_iccv, rangesh2021autonomous}. However, the state-of-the-art for gaze estimation does not account for the complexities of the real world (for example - drivers wearing eyeglasses, harsh illumination, nighttime conditions, etc.). According to the Vision Impact Institute~\cite{visit_essilor_global_website, vision_impact_institute}, one in five drivers have a vision problem and most of them wear correction eyeglasses while driving. Additionally, according to the National Safety Council, 50\% of all accidents happen while driving in the dark~\cite{driving_at_night}. However, it is difficult to estimate the gaze with a high accuracy using RGB images acquired from dark environments. It is primarily because RGB cameras suffer from a lack of photons captured by imaging sensors, resulting in low signal-to-noise ratios~\cite{cameras_Gamadia_2007}. In this study, we find that using infrared cameras can mitigate these lighting issues; however, for face images with eye-wear, data pre-processing or algorithmic improvements are necessary to identify the driver's gaze zones. Since previous works on gaze estimation~\cite{Vora_2018} achieved decent results (under ideal conditions) by using convolutional neural networks (CNNs), we adopt their basic methodology and develop novel pre-processing approaches that can improve gaze estimation in these demanding conditions.

\begin{figure}[t]
    \center{\includegraphics[width=0.47\textwidth]
    {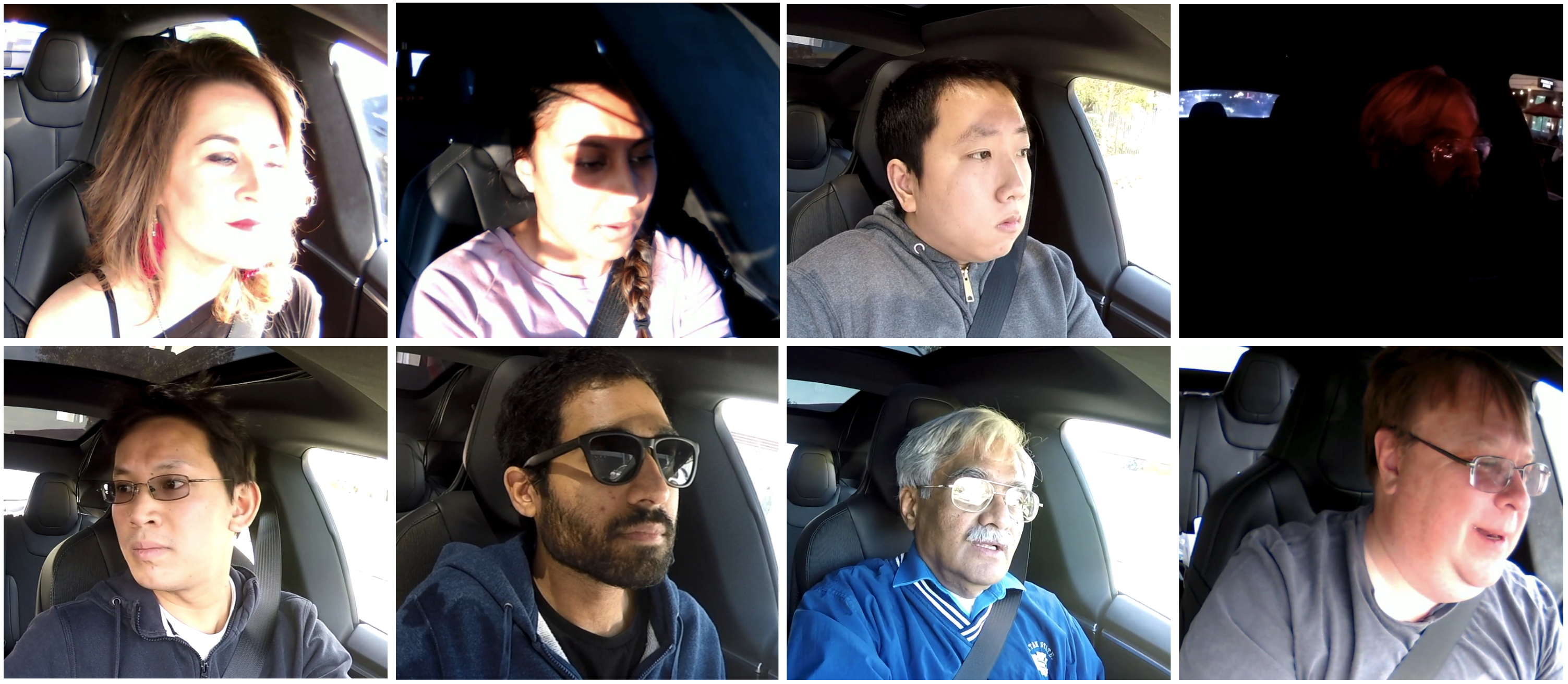}}
    \caption{\label{fig:motivation} The real world introduces variability and complexity that driver gaze estimation systems usually ignore. Some examples include - use of eyeglasses, harsh illumination, nighttime data etc.}
\end{figure}

Generative Adversarial Networks (GANs) \cite{goodfellow2014generative} have shown promising results on a variety of computer vision tasks like image super-resolution~\cite{Super_resolution_Ledig_2017}, realistic face images generation~\cite{StyleGAN_Karras_2019}, style transfer, image editing~\cite{MUNIT_Huang_2018, CycleGAN_Zhu_2017}, etc. There are two sub-models in GAN architecture, a generator and a discriminator. The generator updates itself to synthesize images that are indistinguishable from real images by learning the data distribution. The discriminator's task is to differentiate between real and synthesized images. Both of them are trained together in an adversarial manner until an equilibrium is attained. More recent work on GANs has focused on addressing key issues like training on large datasets, training larger models, improving stability during training, preserving finer details, etc.  In this study, we build on one such model called the CycleGAN~\cite{CycleGAN_Zhu_2017}, which achieved the state-of-the-art on image-to-image translation by using unpaired images from the source and target distributions.
\begin{table*}[t]
\centering
\caption{Related research} \label{tab:related_research_all}\
\resizebox{0.9\linewidth}{!}{%
\begin{threepartsubtable}\centering
\subcaption{Selected research on gaze estimation in vehicular environments}
\begin{tabular}{|c|c|c|c|c|c|}
\hline

\textbf{Study} & \textbf{Objective} & \textbf{Sensor} & \textbf{Features} & \textbf{Capture conditions} & \textbf{Methodology} \\ \hline

Vora et al.~\cite{Vora_2018} & Gaze zone classification using CNNs & 1 RGB camera & HP\tnote{1} \& gaze & \begin{tabular}[c]{@{}c@{}}Daytime; \\ w/o eyeglasses\end{tabular} & CNN \\ \hline

Martin et al.~\cite{Martin_2018} & \begin{tabular}[c]{@{}c@{}}Estimating gaze dynamics,\\  glance duration and frequency\end{tabular} & 1 RGB camera & HP \& gaze & \begin{tabular}[c]{@{}c@{}}Daytime; \\w/o eyeglasses\end{tabular} & CNN \& geometry \\ \hline

Naqvi et al.~\cite{Naqvi_NIR_2018} & \begin{tabular}[c]{@{}c@{}}Gaze zone detection using NIR\tnote{2}\\ camera and deep learning\end{tabular} & \begin{tabular}[c]{@{}c@{}}1 NIR camera \&\\  NIR LEDs\end{tabular} & HP \& gaze & \begin{tabular}[c]{@{}c@{}}Daytime \& nighttime; \\ w \& w/o eyeglasses\end{tabular} & CNN \\ \hline

Yong et al.~\cite{Yoon_2NIR_2019} & \begin{tabular}[c]{@{}c@{}}Gaze detection using dual NIR\\ cameras and deep residual networks\end{tabular} & \begin{tabular}[c]{@{}c@{}}2 NIR cameras \&\\ NIR LEDs\end{tabular} & HP \& gaze & \begin{tabular}[c]{@{}c@{}}Daytime \& nighttime; \\ w/ \& w/o eyeglasses\end{tabular} & CNN \\ \hline

Jha \& Busso~\cite{Jha_2018} & \begin{tabular}[c]{@{}c@{}}Gaze region estimation\\  using dense pixelwise predictions\end{tabular} & \begin{tabular}[c]{@{}c@{}}1 RGB camera \& \\ 1 headband with AprilTags\end{tabular} & HP & \begin{tabular}[c]{@{}c@{}}Daytime;\\ w/ \& w/o eyeglasses\end{tabular} & \begin{tabular}[c]{@{}c@{}}Dense Neural\\  Networks\end{tabular} \\ \hline

Wang et al.~\cite{Wang_2019} & \begin{tabular}[c]{@{}c@{}}Continuous gaze estimation\\ using RGB-D camera\end{tabular} & 1 RGB-D camera & HP \& gaze & \begin{tabular}[c]{@{}c@{}}Daytime;\\ w/ \& w/o eyeglasses\end{tabular} & \begin{tabular}[c]{@{}c@{}}Feature extraction\\ \& k-NN\end{tabular}\\ \hline

\end{tabular}
\label{tab:related_research_gaze}
\end{threepartsubtable}
}

\bigskip

\resizebox{0.99\linewidth}{!}{%
\begin{threepartsubtable}\centering
\subcaption{Selected research on eyeglass removal and related topics}%
\begin{tabular}{|c|c|c|c|c|c|}
\hline

\textbf{Study} & \textbf{Objective} & \textbf{Methodology} & \textbf{Dataset} & \textbf{Advantages} & \textbf{Disadvantages} \\ \hline

Zhu et al.~\cite{CycleGAN_Zhu_2017} & \begin{tabular}[c]{@{}c@{}}Unpaired image-to-image\\ translation\end{tabular} & Cycle-GAN\tnote{3} & \begin{tabular}[c]{@{}c@{}}Unpaired images\end{tabular} & \begin{tabular}[c]{@{}c@{}}Uses unpaired images;\\performs well on style transfer\end{tabular} & \begin{tabular}[c]{@{}c@{}}Not realistic for eyeglass removal;\\ does not preserve gaze direction\end{tabular} \\ \hline

Hu et al.~\cite{hu2019unsupervised} & \begin{tabular}[c]{@{}c@{}}Eyeglass removal\end{tabular} & ER-GAN & \begin{tabular}[c]{@{}c@{}}Unpaired \\ CelebA \& LFW\end{tabular} & \begin{tabular}[c]{@{}c@{}}Good performance on eyeglass\\ removal for frontal faces\end{tabular} & \begin{tabular}[c]{@{}c@{}}Gaze is not preserved; only works\\ for aligned frontal images\end{tabular} \\ \hline

Amodio et al.~\cite{hu2019unsupervised} & \begin{tabular}[c]{@{}c@{}} Unpaired image-to-image\\ translation \end{tabular} & TraVeLGAN & \begin{tabular}[c]{@{}c@{}}Unpaired Images\\from multiple domains\end{tabular} & \begin{tabular}[c]{@{}c@{}}Good performance on style\\ transfer and eyeglass removal\end{tabular} & \begin{tabular}[c]{@{}c@{}}Eyes are often swapped; gaze is \\not preserved\end{tabular} \\ \hline

Wang et al.~\cite{wang_ECGAN_2018} & \begin{tabular}[c]{@{}c@{}}Facial obstruction\\  removal\end{tabular} & \begin{tabular}[c]{@{}c@{}}EC-GAN\\ \& LS-GAN\end{tabular} & Paired CelebA & \begin{tabular}[c]{@{}c@{}}Better results than Cycle-GAN;\\ improved face recognition accuracy\end{tabular} & Needs an obstruction classifier \\ \hline

Liang et al.~\cite{Liang2017DeepCN} & \begin{tabular}[c]{@{}c@{}}Learn mappings between faces\\ with and without glasses\end{tabular} & CNN & \begin{tabular}[c]{@{}c@{}}Images from web \&\\surveillance cameras\end{tabular} & \begin{tabular}[c]{@{}c@{}}Single step, end-to-end method\end{tabular} & \begin{tabular}[c]{@{}c@{}}Images need to be aligned; needs paired\\ images; uses synthesized eyeglasses\end{tabular} \\ \hline

Li et al.~\cite{li2016deep} &  \begin{tabular}[c]{@{}c@{}}Identity-aware transference\\of facial attributes\end{tabular} & DIAT-GAN & \begin{tabular}[c]{@{}c@{}}Aligned CelebA\end{tabular} & \begin{tabular}[c]{@{}c@{}}Flexibility to modify different\\ facial attributes\end{tabular} & \begin{tabular}[c]{@{}c@{}}Generated gaze is different;\\ blurry outputs\end{tabular} \\ \hline

Shen et al.~\cite{Shen_2017} & \begin{tabular}[c]{@{}c@{}}Facial attribute\\manipulation\end{tabular} & \begin{tabular}[c]{@{}c@{}}GAN\end{tabular} & \begin{tabular}[c]{@{}c@{}}CelebA \&  LFW\end{tabular} & \begin{tabular}[c]{@{}c@{}}Capability to manipulate images\\ with modest pixel modifications\end{tabular} & \begin{tabular}[c]{@{}c@{}}Does not completely remove eyeglasses;\\the eye region is not preserved
\end{tabular}\\ \hline
\end{tabular}
 \begin{tablenotes}
    \item[1] Head Pose
    \item[2] Near Infra-Red
    \item[3] Generative Adversarial Networks
 \end{tablenotes}
  \label{tab:related_research_glasses_removal}
 \end{threepartsubtable}
}
\end{table*}

Inspired by the CycleGAN architecture, we propose the Gaze Preserving CycleGAN (GPCycleGAN) which makes use of an additional gaze consistency loss. GPCycleGAN has the following advantages compared to other gaze estimation methods and glass removal techniques: First, GPCycleGAN preserves the gaze and allows for more accurate gaze estimation on images with eyeglasses. Second, unlike previous works~\cite{Liang2017DeepCN, wang_ECGAN_2018} on eyeglass removal, there is no need for paired images to train GPCycleGAN. Third, it works in different environments, lighting conditions, eyeglass types, and with significant variations of the head pose.

The four main contributions of this paper are: a) An in-depth analysis of traditional gaze estimation under different conditions (e.g., with and without eyeglasses, daytime, nighttime, etc.), and its shortcomings, b) The GPCycleGAN model for eyeglass removal specifically optimized for the gaze classification task, c) Experimental analyses to illustrate how the GPCycleGAN model improves the accuracy, interpretability and robustness over a variety of baseline models, and d) A naturalistic driving dataset with labeled gaze zones that includes both IR and RGB images. This work is an extension of our research presented in~\cite{rangesh2020driver}. In this study, we extend and refine our previous approach with new losses and training schemes, update our previously reported quantitative numbers, and provide new qualitative analysis.

\section{Related Research}
Recent related studies are listed in Table~\ref{tab:related_research_all} with 2 sub-tables that represent two research areas - gaze estimation and eyeglass removal. We mainly focus on vision-based approaches post 2016 for gaze estimation. Please refer to the following studies for earlier work: Vora et al.~\cite{Vora_2018} for gaze estimation; Kar et al.~\cite{Kar_2017} for eye-gaze estimation systems, algorithms, and performance evaluation methods in consumer platforms; a survey on driver behavior analysis for safe driving by Kaplan et al.~\cite{7225158};  a review on driver inattention monitoring systems by Dong et al.~\cite{5665773}; and a survey on head pose estimation by Murphy et al.~\cite{10.1109/TPAMI.2008.106}. For the second research area, we were unable to find published studies focused on eyeglass removal for drivers’ face images. We instead provide discussions on general-purpose eyeglass removal papers.

\subsection{Gaze Estimation}
 Gaze estimation studies can be categorized in the following ways. First, by the type of model used, they can be divided into convolutional neural networks\cite{Vora_2018}, statistical learning models \cite{Fridman_2016}, or geometric approaches\cite{Martin_2018}. Second, methods can be distinguished by the cues they consider, i.e. studies that only use head pose~\cite{Jha_2018} versus ones that use both head pose and eye information \cite{Yoon_2NIR_2019, Vora_2018,Tawari2014Where, OhnBar2014head}. Third, research can be categorized by the conditions they capture. For instance, studies with limited illumination changes~\cite{Vora_2018} versus research with multiple environments and the variations that come with it~\cite{Yoon_2NIR_2019}. Lastly, studies can be separated by the sensors they use, including RGB cameras~\cite{Vora_2018}, IR cameras~\cite{chinsatit2017cnn}, NIR cameras~\cite{Naqvi_NIR_2018}, multiple cameras~\cite{TawariGazeEstimationGlasses2014}, and wearable sensors~\cite{Tsukada_2011}. However, only a few studies emphasize the problems associated with driving with eyeglasses or eye-wear. Tawari et al.~\cite{TawariGazeEstimationGlasses2014} and Lee et al.~\cite{LeeGazeGlasses2011} mention the unreliability of gaze estimation for drivers with glasses, and propose methods that only rely on head pose as fallback solutions. Naqvi et al.~\cite{Naqvi_NIR_2018} combine head pose estimation and pupil detection to determine the eye gaze, but their method only works under ideal capture conditions. Jah and Busso~\cite{Jha_2018} use only head pose in their method. Wang et al. \cite{Wang_2019} combine depth images of the head and RGB images of eyes, but gaze estimation for eye images is unstable and only works with frontal images under ideal conditions. In~\cite{Yoon_2NIR_2019}, Yoon et al. collect a dataset comprising of images in daytime/nighttime, images with and without eyeglasses using two NIR cameras and NIR lights. Although they achieve good performance, they do not explicitly model the presence of eyeglasses in images. In this study, we show that such approaches tend not to generalize to different settings and usually overfit the training set.

\subsection{Eyeglass Removal}
In real-world cabin environment, large variations, including head pose, eyeglass type, environment conditions, and the presence of reflection artifacts, make eyeglass removal is a complex task. Existing eyeglass removal studies make use of statistical learning, principal component analysis (PCA), or deep learning, including GAN based methods. Statistical learning and PCA were the primary approaches prior to the advent of deep learning. These approaches do not require high computational power but have limitations on eyeglasses, environmental conditions, and head poses~\cite{GR_Yi_2011, GR_Wong_2013}. On the other hand, methods using deep learning, such as ones proposed by Liang et al.~\cite{Liang2017DeepCN}, Wang et al.~\cite{wang_ECGAN_2018}, Din et al.~\cite{Din2020Face}, and Lee et al.~\cite{Lee2020ByeGlassesGAN} modified GANs for better results, but they need paired and aligned images for training. Such datasets are expensive and tedious to collect. Different GAN architectures proposed in~\cite{CycleGAN_Zhu_2017},~\cite{hu2019unsupervised}, and\cite{TraVeLGAN_Amodio_2019} do not require paired images, but their models fail to keep gaze information and do not perform well on non-frontal images. Models developed by Li et al.~\cite{li2016deep} and Shen et al.~\cite{Shen_2017} are capable of changing multiple facial attributes but do not produce satisfying results on eyeglass removal. All the above methods on eyeglass removal have general or different purposes. However, none of these methods are constructed to preserve gaze of face images, and hence cannot necessarily be used out-of-the-box for gaze estimation.

\section{Dataset}
\begin{figure}[t]
    \center{\includegraphics[width=0.48\textwidth]
    {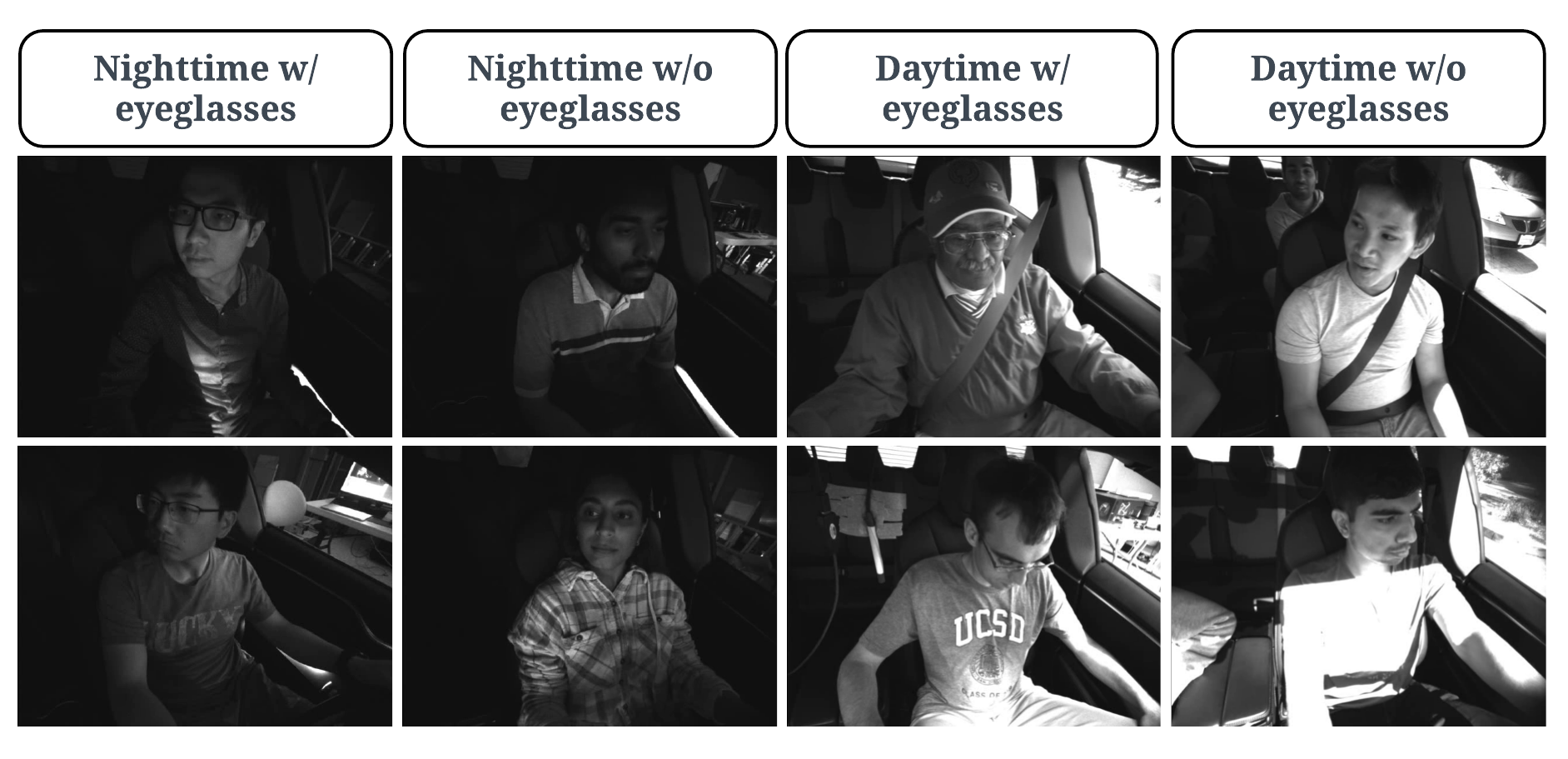}}
    \caption{\label{fig:dataset} Example images from our dataset under different capture conditions.}
\end{figure}

Since our primary goal is to design a gaze estimation system for the real-world, we prioritized using small form-factor infrared cameras with suitable real-time performance. We decided on an Intel RealSense IR camera and enclosed it in a custom 3D printed enclosure mounted next to the rearview mirror. To ensure a good compromise between larger fields-of-view and faster processing speeds, we settled on a capture resolution of $640 \times 480$. Similar to Vora et al.~\cite{Vora_2018}, we divide the driver's gaze into seven gaze zones: \textit{Eyes Closed/Lap}, \textit{Forward}, \textit{Left Mirror}, \textit{Speedometer}, \textit{Radio}, \textit{Rearview}, \textit{Right Mirror}. Unlike previous studies, we also include gaze zones related to driver inattention or unsafe driving behavior. Our entire dataset comprises of 13 subjects in different lighting conditions (daytime, nighttime and harsh lighting), wearing a variety of eyeglasses. For every gaze zone, participants were instructed to keep their gaze fixed while moving their heads within reasonable limits. Of the 13 subjects, 7 were male, with an ages ranging from 20 to 65 years. In total, 336177 frames of images were captured, which we split into training, validation, and test sets with no overlap of subjects. Table~\ref{tab:dataset} shows the distribution of images and subjects across different splits and capture conditions. Fig.~\ref{fig:dataset} depicts exemplar images from our dataset. We ensure that the dataset is suitably diverse and challenging to represent the complexities observed in the real world. 

\begin{table}[t]
\begin{center}
\caption{\label{tab:dataset}Dataset size (\# of images, \# of subjects) across different splits and capture conditions. There is no overlap of subjects between different splits to ensure cross-subject validation and testing.}
\begin{tabular}{|c | c | c | c | c |} 
\hline
\backslashbox[35mm]{\textbf{\shortstack{Capture\\conditions}}}{\textbf{\shortstack{    \\Dataset\\split}}} & \textbf{Training} & \textbf{Validation} & \textbf{Testing} \\ [0.5ex]\hline\hline
\textbf{daytime; w/o eyeglasses} & (67151, 9)  & (9908, 1) & (2758, 4)\\ \hline
\textbf{nighttime; w/o eyeglasses} & (59352, 9) & (8510, 1) & (2768, 4)\\ \hline
\textbf{daytime; w/ eyeglasses} & (43432, 5) & (9062, 1) & (3294, 4)\\ \hline
\textbf{nighttime; w/ eyeglasses} & (33189, 5) & (8103, 1) & (2897, 4)\\ \hline\hline
\textbf{Total (all conditions)} & (203124, 9) & (35583, 1) & (11717, 4)\\ \hline
\end{tabular}
\end{center}
\end{table}

\begin{figure}[t]
    \center{\includegraphics[width=0.35\textwidth]
    {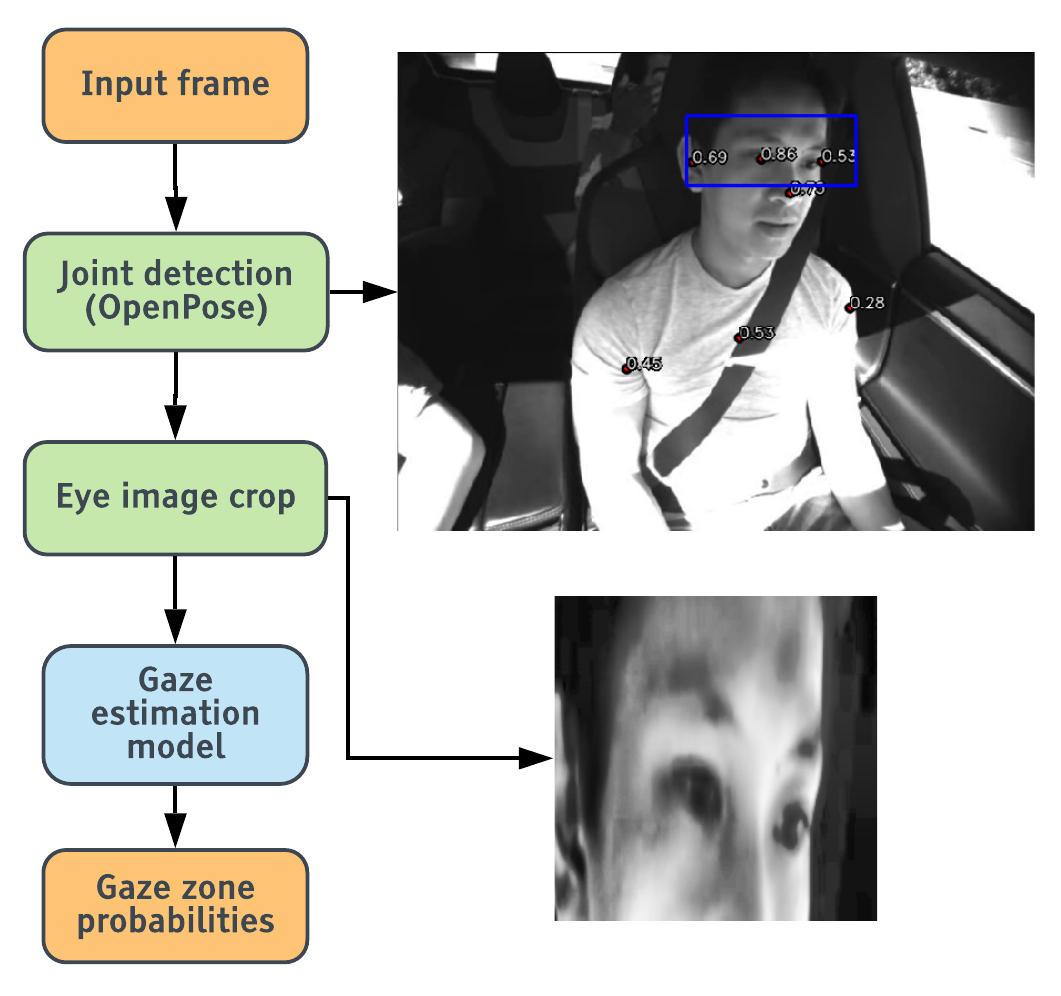}}
    \caption{\label{fig:pipeline} Overall processing pipeline for driver gaze zone estimation.}
\end{figure}

\begin{table*}[t]
\begin{center}
\caption{\label{tab:val}Validation accuracies for gaze models trained on data with different capture conditions}
\resizebox{0.95\linewidth}{!}{%
\begin{tabular}{| l | c | c | c | c | c | c | c | c | c |} 
\hline
\backslashbox[40mm]{\textbf{\shortstack{Model \circled{\#} \& \\training data used}}}{\textbf{\shortstack{ \\ \\Validation\\data }}} & \textbf{\shortstack{\circled{a} daytime;\\w/o eyeglasses}} & \textbf{\shortstack{\circled{b} nighttime;\\w/o eyeglasses}} & \textbf{\shortstack{\circled{c} daytime;\\w/ eyeglasses}} & \textbf{\shortstack{\circled{d} nighttime;\\w/ eyeglasses}} & \textbf{ \shortstack{\circled{e} w/o\\ eyeglasses}} & \textbf{\shortstack{\circled{f} w/\\ eyeglasses}} & \textbf{\shortstack{\textcircled{g}\\ daytime}} & \textbf{\shortstack{\circled{h}\\ nighttime}} & \textbf{ \shortstack{\circled{i} all\\conditions}}\\ [0.5ex]\hline

\textbf{\circled{1} daytime; w/o eyeglasses}  & 81.0774  & 62.4657 &  45.4588  & 11.0680&   72.5677  & 29.7526 &  64.1875 &  38.1209 &  52.2757 \\ 
\textbf{\circled{2} nighttime; w/o eyeglasses} & 73.5272 &  87.2226 &  29.3782  & 25.6684 &  79.7891 &  27.6839  & 52.5923  & 58.0671  & 55.0941 \\ 
\textbf{\circled{3} daytime; w/ eyeglasses} & 60.4746 & 42.7325  & 70.6334 &  39.7423&   52.3625  & 56.5256 &  65.2918  & 41.3162 &  54.3356 \\ 
\textbf{\circled{4} nighttime; w/ eyeglasses} & 51.4649  & 76.7639 &  45.3773  & 64.8151 &  63.0322 &  54.2544 &  48.5782  & 71.1043 &  58.8720 \\ 
\textbf{\circled{5} w/o eyeglasses} &  85.5508  & 88.7809 &  43.9101  & 20.6954 &  87.0276  & 33.3080  & 65.8053  & 56.5317  & 61.5675\\ 
\textbf{\circled{6} w/ eyeglasses} & 71.3746  & 82.2239 &  69.5738  & 77.7254 &  76.3351 &  73.2966 &  70.5207 &  80.0932&   74.8951\\ 
\textbf{\circled{7} daytime} & 77.2131  & 74.7569 & 73.5095  & 39.9224 &  76.0901 &  58.1704  & 75.4569 &  58.2573 &  67.5971\\ 
\textbf{\circled{8} nighttime} & 55.1192 &  83.7696  & 49.3596   &61.3520  & 68.2189 &  54.8365 &  52.3881  & 73.1514 &  61.8763 \\ 
\textbf{\circled{9} all conditions} & 81.6969 &  83.3084  & 56.7885  & 66.4358 &  82.4337  & 61.1944  & 69.8857 &  75.3166 &  72.3675\\ \hline

\end{tabular}
}
\end{center}
\end{table*}

\section{Methodology}
As depicted in Fig.~\ref{fig:pipeline}, the steps involved in our proposed gaze estimation pipeline are as follows: (a) landmark detection using OpenPose~\cite{cao2018openpose}, (b) eye image cropping, resizing, and equalization, (c) gaze estimation. We use OpenPose for landmark detection because of its high accuracy under different conditions and fast inference speed. Based on the work by Vora et al.~\cite{Vora_2018}, we crop the eye region using the estimated landmarks as per their conclusion that the upper half of the face as an input produced the best results for downstream gaze classification. Next, we use adaptive histogram equalization to improve the contrast and resize the images to $256 \times 256$ before feeding them to the gaze estimation models.

\subsection{Issues with Lighting \& Eyeglasses}
To understand the impacts of different conditions on gaze estimation, we carry out an extensive experiment to analyze the performance of models trained with subsets of data, when they are tested on data from within and outside the training distribution.
Table~\ref{tab:val} shows validation accuracies of SqueezeNet-based gaze classifier models \circled{1} - \circled{9} (as proposed in~\cite{Vora_2018}), each trained on data captured under different conditions \circled{a} - \circled{i}. From the Table, we glean that model \circled{5} validated on data \circled{a}, \circled{b}, and \circled{e} produce similar accuracies, and model \circled{9} validated on data \textcircled{g}, \circled{h}, and \circled{i} also perform similarly. This demonstrates that the gaze classifier models work well on daytime and nighttime data when trained on data containing both conditions. Thus, problems related to lighting can be effectively solved by training using IR images, appropriate normalization, and histogram equalization. We also learn that nighttime data is easier to model and results in better accuracies in general. Next, to analyze the models' performance on data with and without glasses, we observe that the accuracy of model \circled{5} validated on data \circled{e} (87.0276\%) is much higher than the accuracy of model \circled{6} validated on data \circled{f} (73.2966\%). Similarly, the accuracy of model \circled{1} is better than model \circled{3}, and model \circled{2} is better than model \circled{4} when their validation sets comprise of data from their training distribution. The model trained with all the data (\circled{9}) always has better accuracies when validated on data without glasses (\circled{a}, \circled{b}, and \circled{e}), than that with glasses (\circled{c}, \circled{d}, and \circled{f}). This implies that simply training on data with and without eyeglasses is not sufficient to ensure good generalization across both conditions. Thus, we need to explicitly handle eyeglasses through modeling, or reduce the \textit{domain gap} between images with and without eyeglasses. In this study, we propose to do the latter by training an eyeglass removal network.
\begin{figure*}[ht]
    \center{\includegraphics[width=0.95\textwidth]
    {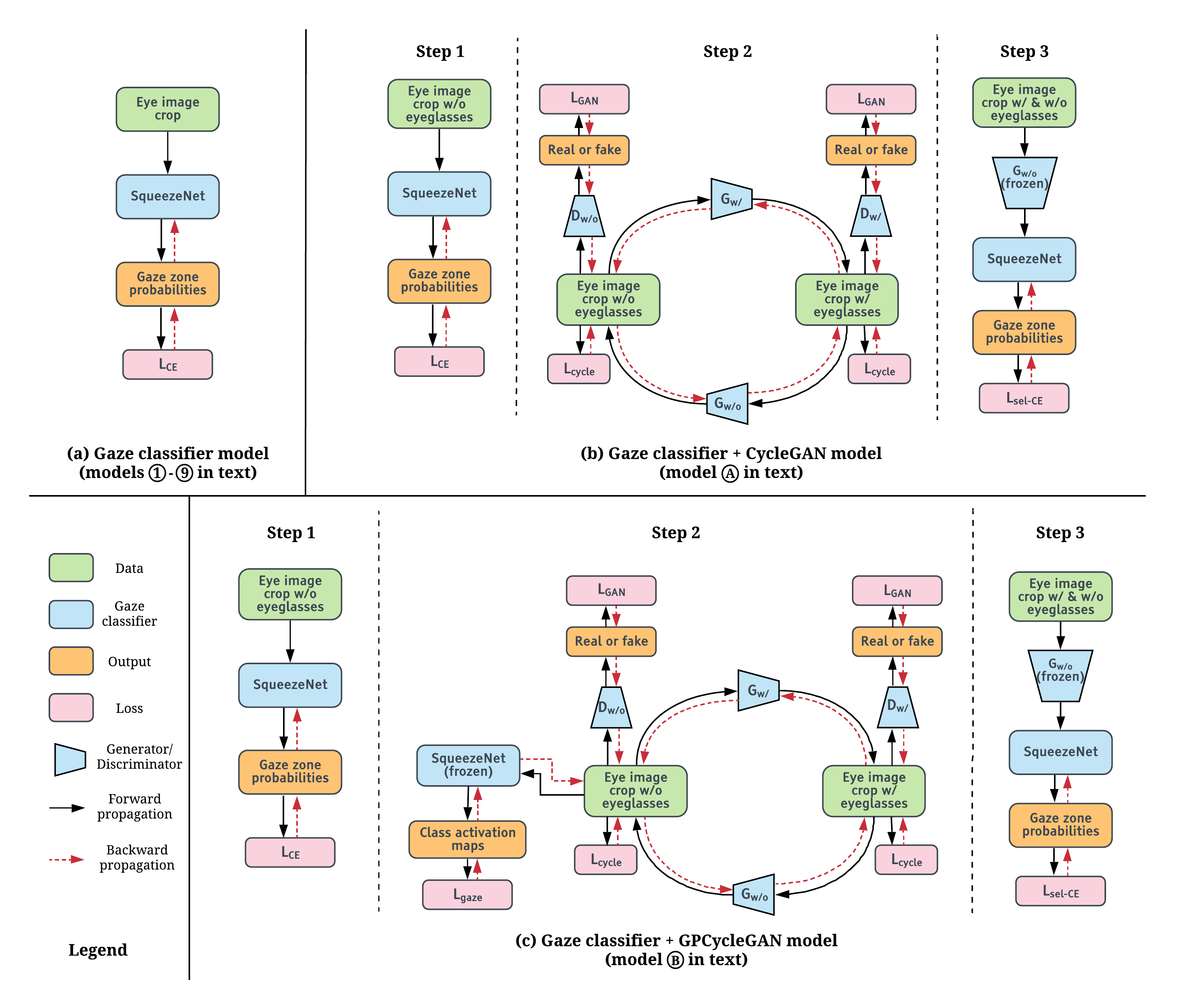}}
    \caption{\label{fig:training} Training setup and architectures for different gaze zone estimation models.}
\end{figure*}

\begin{figure}[ht]
    \center{\includegraphics[width=0.35\textwidth]
    {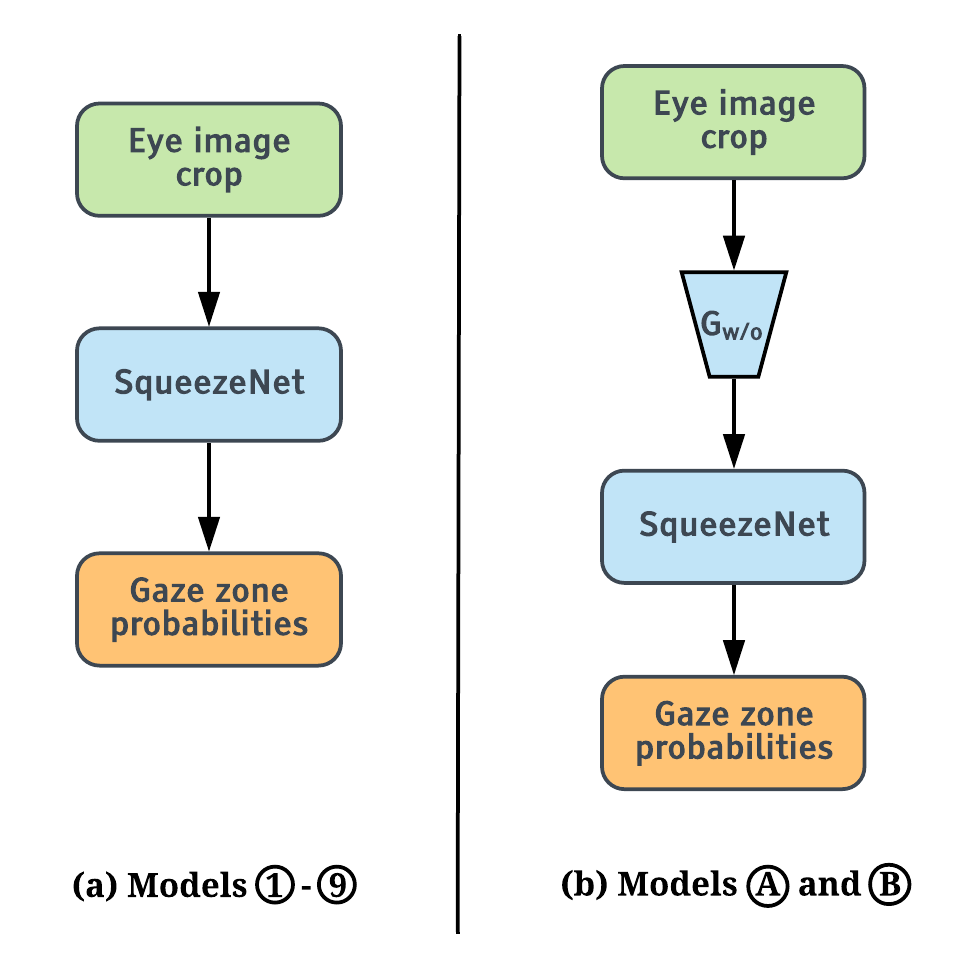}}
    \caption{\label{fig:inference} Inference setup and architectures for different gaze zone estimation models.}
\end{figure}

\subsection{Vanilla CycleGAN-based Approach}

The two domains for the eyeglass removal problem are eye image crops without eyeglasses, $\mathcal{X}$, and eye image crops with eyeglasses, $\mathcal{Y}$ ($\mathcal{X}, \mathcal{Y} \subset \mathbb{R}^{H \times W \times C}$, $H = 256$, $W = 256$, and $C = 1$ for IR images, $C = 3$ for RGB images). We denote $X\ (X\in \mathcal{X})$ and $Y\ (Y \in \mathcal{Y})$ as sample images from domains $\mathcal{X}$ and $\mathcal{Y}$ respectively. 

First, we consider a baseline gaze classifier model (Step 1 in Fig. \ref{fig:training}(b)), for which we use input images from domain $\mathcal{X}$. We use the SqueezeNet architecture as before and denote its output gaze zone probabilities as $\{p_{i}\}_i$. Standard cross-entropy loss ($\mathcal{L}_{CE}$) is used for training the gaze classifier. $\mathcal{L}_{CE}$ is defined as:
\begin{equation} \label{CE_loss}
\mathcal{L}_{CE} = -\sum_{i} z_{i} \log p_{i},
\end{equation}
where $i$ is the class index and $z$ is the one-hot encoding of the ground truth class.

Next, we consider a CycleGAN-based eyeglass removal network, followed by a SqueezeNet-based gaze classifier (Step 2 in Fig.~\ref{fig:training}(b)). The CycleGAN-based eyeglass removal model learns the mapping $G_{w/o}$ from $\mathcal{Y}$ to $\mathcal{X}$ (removing eyeglasses), denoted as $G_{w/o}$: $\mathcal{Y} \rightarrow \mathcal{X}$. When trained, $G_{w/o}$ generates images $G_{w/o}(\cdot)$ using the input from either domain, and then passes them to the pre-trained gaze classifier to generate output probabilities $\{p_i\}_i$. 
In a similar manner, the mapping for adding eyeglasses to $X \in \mathcal{X}$ is $G_{w/}: \mathcal{X} \rightarrow \mathcal{Y}$, and it is trained simultaneously with $G_{w/o}$. The discriminator $D_{w/}$ aims to distinguish between real images $Y$ and generated images $G_{w/}(X)$, whereas the discriminator $D_{w/o}$ aims to do so between real images $X$ and generated images $G_{w/o}(Y)$. The losses used in this model are: cycle consistency losses \cite{CycleGAN_Zhu_2017} (Eq.~\ref{eq:cycle_loss}) that encourages cycle consistency between the real images and the reconstructed images; adversarial losses \cite{goodfellow2014generative} (Eq.~\ref{eq:GAN_loss}) for making the generated images indistinguishable from real images; and identity losses \cite{taigman2016unsupervised} (Eq.~\ref{eq:identity_loss}) to ensure identity mapping when the input belongs to the target domain. 

\begin{equation} \label{eq:cycle_loss}
\begin{aligned}
\mathcal{L}_{\text{cyc}}(G_{w/}, G_{w/o}) &= \mathbb{E}_{X \sim P(X)}\left[\|G_{w/o}(G_{w/}(X))-X\|_{1}\right] \\
& + \mathbb{E}_{Y \sim P(Y)}\left[\|G_{w/}(G_{w/o}(Y))-Y\|_{1}\right],
\end{aligned}
\end{equation}
\begin{equation} \label{eq:GAN_loss}
\begin{aligned}
\mathcal{L}_{\text{adv}} & (G_{w/}, G_{w/o}, D_{w/}, D_{w/o}) & \\
& = \mathbb{E}_{Y \sim P(Y)}\left[\log D_{w/}(Y)\right]\\
& + \mathbb{E}_{X \sim P(X)}\left[\log \left(1-D_{w/}\left(G_{w/}\left(X\right)\right)\right)\right]\\
& + \mathbb{E}_{X \sim P(X)}\left[\log D_{w/o}(X)\right]\\
& + \mathbb{E}_{Y \sim P(Y)}\left[\log \left(1-D_{w/o}\left(G_{w/o}\left(Y\right)\right)\right)\right],
\end{aligned}
\end{equation}
and,
\begin{equation} \label{eq:identity_loss}
\begin{aligned}
\mathcal{L}_{\text{identity }}(G_{w/}, G_{w/o}) &= \mathbb{E}_{Y \sim P(Y)}\left[\|G_{w/}(Y)-Y\|_{1}\right]\\
& + \mathbb{E}_{X \sim P(X)}\left[\|G_{w/o}(X)-X\|_{1}\right],
\end{aligned}
\end{equation}
where $X \sim P(X)$ and $Y \sim P(Y)$ are the image distributions.

The full objective for the CycleGAN model is the following:
\begin{equation} \label{eq:all_CycleGan_loss}
\begin{aligned}
\mathcal{L}_{total} &= \mathcal{L}_{\text{adv}} + \lambda_{1} \mathcal{L}_{\text{cyc}} + \lambda_{2} \mathcal{L}_{\text{identity}},
\end{aligned}
\end{equation}
where $\lambda_{1}$ is the weight of the cycle loss, and $\lambda_{2}$ is the weight of the identity loss. 

As a final refinement step, we retrain the gaze classifier from step 1 using real images from $\mathcal{X}$ and fake images from $G_{w/o}(\mathcal{Y})$ (Step 3 in Fig.\ref{fig:training}(b)). Unlike step 1, we use a \textit{selective} cross-entropy loss ($L_{sel-CE}$) to avoid overfitting to errors resulting from $G_{w/o}(\cdot)$. $\mathcal{L}_{sel-CE}$ is defined as:
\begin{equation}\label{sel_CE_loss}
\mathcal{L}_{sel-CE} = 
\begin{cases}
-\sum_{i} z_{i} \log p_{i}, & \text{if } \argmax_i p_i = \argmax_i z_i\\
0, & \text{otherwise,}
\end{cases}
\end{equation}
where only the cross-entropy loss from correctly classified samples are backpropagated. This ensures that the gaze classifier is not trained to incorrectly classify fake images with a potentially different gaze zone than the original. After the fine-tuning step, the generator $G_{w/o}$ and the gaze classifier are deployed sequentially to estimate gaze zones in the presence of eyeglasses (Fig.~\ref{fig:inference}(b)).  


\subsection{Proposed GPCycleGAN-based Approach}

Our proposed GPCycleGAN model with the gaze classifier (Fig.~\ref{fig:training}(c)) builds on the previous model, with the addition of a gaze consistency loss ($\mathcal{L}_{gaze}$) in step 2. To preserve the gaze features during eyeglass removal, we use the trained gaze classifier from step 1 and compute the gaze consistency loss using its Class Activation Maps (CAMs). The proposed gaze consistency loss is defined as follows:
\begin{equation} \label{gaze_loss}
    \mathcal{L}_{\text {gaze}}(G_{w/}, G_{w/o}) = \frac{1}{N}\sum_{i=1}^{N} ||T(A_{w/o; real}^{i}) - T(A_{w/o; rec}^{i})||_2,
\end{equation}
where $N$ stands for the total number of classes (gaze zones), $i$ is the class index, and $\{A^i\}_{i=1}^{N}$ are the CAMs from the trained gaze classifier. Subscript $w/o; real$ denotes CAMs obtained from a real eye image crop without eyeglasses, and subscript $w/o; rec$ denotes CAMs corresponding to reconstructed images $G_{w/o}(G_{w/}(\cdot))$ using the same eye image crop. 
The transformation $T(\cdot)$ applied to the CAMs is a simple sigmoid function with temperature $\tau$:
\begin{equation} \label{transformation}
    T(A^{i}) = \frac{1}{1 + exp(-\tau \cdot A^{i})}.
\end{equation}
The nature of the proposed gaze loss necessitates the use of a cyclic structure, as we do not have perfectly paired samples of images with and without glasses to enforce gaze consistency directly.

The full objective for the proposed GPCycleGAN model is:
\begin{equation} \label{eq:all_GPCycleGan_loss}
\begin{aligned}
\mathcal{L}_{total} &=\mathcal{L}_{\text{adv}} + \lambda_{1} \mathcal{L}_{\text{cyc}} + \lambda_{2} \mathcal{L}_{\text{identity}} + \lambda_{3} \mathcal{L}_{\text{gaze}},
\end{aligned}
\end{equation}
where $\lambda_{3}$ is the weight for the gaze consistency loss.

\subsection{Implementation Details}
The training and inference architectures are the same for the baseline gaze classification models \circled{1}-\circled{9}, but different for models \circled{A} and \circled{B}. Models \circled{1}-\circled{9} are adopted from Vora et al.~\cite{Vora_2018}, and differ in the data they were trained on (see Table~\ref{tab:val}). Models \circled{A} and \circled{B} are identical in terms of their architectures, and only differ in their training setup and losses (i.e. the addition of a gaze consistency loss in model \circled{B}). Both models consist of generators with 9 residual blocks and $70 \times 70$ PatchGANs~\cite{CycleGAN_Zhu_2017} as discriminators. As illustrated in Fig.~\ref{fig:training}, training models \circled{A} and \circled{B} proceeds in 3 steps: Step 1 involves training a SqueezeNet gaze classifier using eye image crops from domain $X$; Step 2 is for training the generator $G_{w/o}$ in CycleGAN/GPCycleGAN; Step 3 is to fine-tune the gaze classifier from step 1 using generated images $G_{w/o}(x, y)$. The inference for models \circled{1} - \circled{9} (Fig.~\ref{fig:inference}(a)) is a simple forward propagation through the gaze classifier to obtain the gaze zone probabilities. Models \circled{A} and \circled{B} have the same inference setup (Fig.~\ref{fig:inference}(b)), where eye image crops are first passed to the generator $G_{w/o}$ for eyeglass removal, after which they are fed to the gaze classifier which outputs the gaze zone probabilities. 

We choose $\tau = 0.01$ in Eq.~\ref{transformation}, and $\lambda_1 = 10$, $\lambda_2 = 5$, $\lambda_3 = 1$ in Eq.~\ref{eq:all_CycleGan_loss} and Eq.~\ref{eq:all_GPCycleGan_loss}. We use a learning rate of $0.0004$ with an Adam optimizer for training all SqueezeNet classifiers, a learning rate of $0.0002$ for CycleGAN/GPCycleGAN in step 2, and a reduced learning rate of $0.0001$ for the final finetuning step. The gaze classifiers are trained for a total of $50$ epochs, while the GANs are trained for $15$ epochs - both with early stopping. The total runtime for inference using our approach is 27 ms when using an RTX 2080 graphics card (OpenPose: 22 ms, eyeglass removal network: 3 ms, and gaze classifier network: 2 ms) - indicating real-time performance.


\begin{table}[t]
\begin{center}
\caption{\label{tab:test}Test set metrics for different models}
\resizebox{0.95\linewidth}{!}{%
\begin{tabular}{|c | c | c | c |} 
\hline
\textbf{Model} & \textbf{\shortstack{Micro-average\\accuracy(\%)}} & \textbf{\shortstack{Macro-average\\accuracy(\%)}} & \textbf{\shortstack{Confusion\\matrix}} \\ \hline\hline
\circled{5} & 65.27 & 62.13 & Fig.~\ref{fig:cm5}\\ \hline
\circled{9} & 76.77 & 75.27 & Fig.~\ref{fig:cm9}\\ \hline
\circled{A} & 73.76 & 71.31 & Fig.~\ref{fig:cma}\\ \hline
\circled{A} with fine-tuning & 78.16 & 78.01 & Fig.~\ref{fig:cmaft}\\ \hline
\circled{B} & 73.91 & 71.88 & Fig.~\ref{fig:cmb}\\ \hline
\circled{B} with fine-tuning & \textbf{81.23} & \textbf{79.44} & Fig.~\ref{fig:cmbft}\\ \hline
\end{tabular}
}
\end{center}
\end{table}

\begin{figure}[t]
\begin{center}
\begin{subfigure}{.48\linewidth}
  \centering
  \includegraphics[width=1\linewidth]{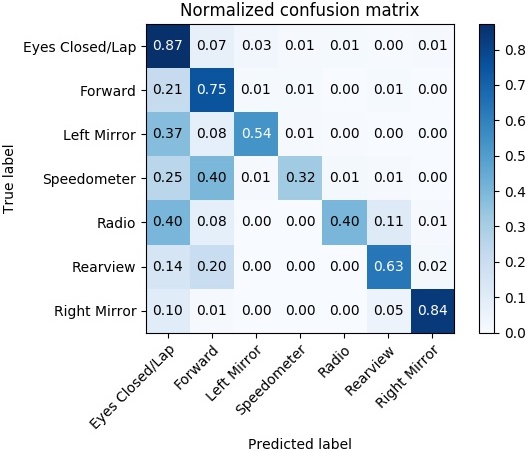}  
  \caption{Model \circled{5}}
  \label{fig:cm5}
\end{subfigure}
\begin{subfigure}{.48\linewidth}
  \centering
  \includegraphics[width=1\linewidth]{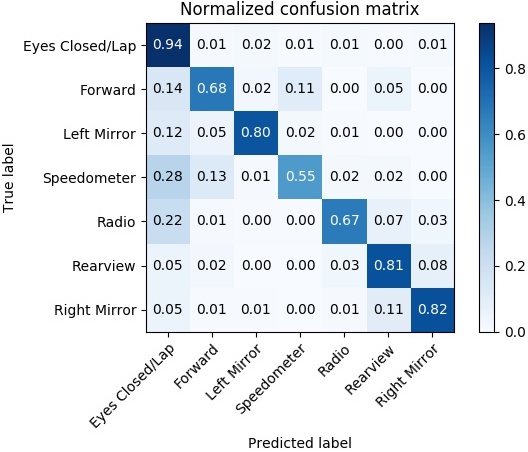}  
  \caption{Model \circled{9}}
  \label{fig:cm9}
\end{subfigure}
\begin{subfigure}{.48\linewidth}
  \centering
  \includegraphics[width=1\linewidth]{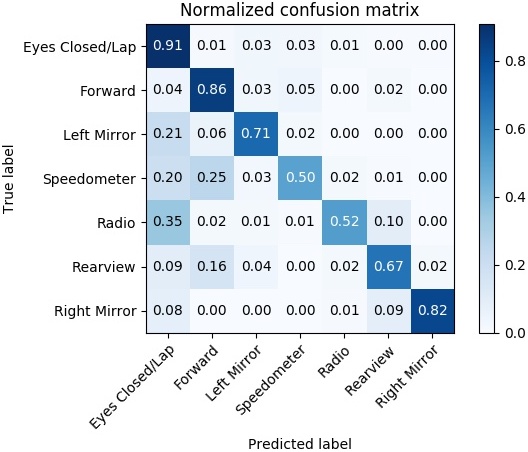}  
  \caption{Model \circled{A}}
  \label{fig:cma}
\end{subfigure}
\begin{subfigure}{.48\linewidth}
  \centering
  \includegraphics[width=1\linewidth]{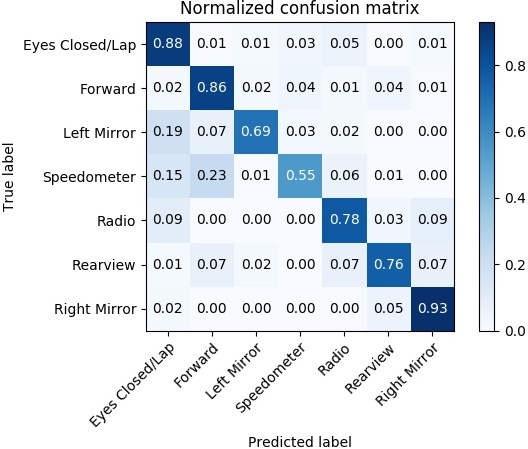}  
  \caption{Model \circled{A} with fine-tuning}
  \label{fig:cmaft}
\end{subfigure}
\begin{subfigure}{.48\linewidth}
  \centering
  \includegraphics[width=1\linewidth]{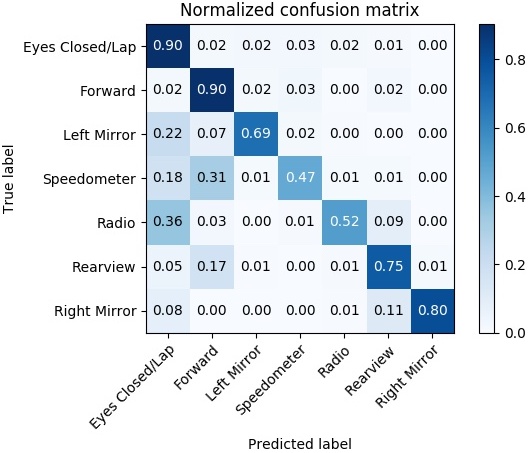}  
  \caption{Model \circled{B}}
  \label{fig:cmb}
\end{subfigure}
\begin{subfigure}{.48\linewidth}
  \centering
  \includegraphics[width=1\linewidth]{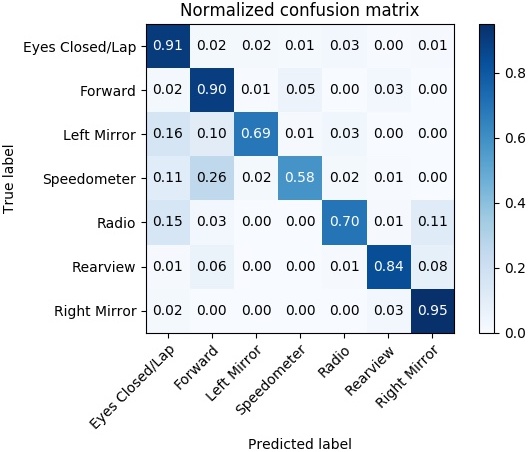}  
  \caption{Model \circled{B} with fine-tuning}
  \label{fig:cmbft}
\end{subfigure}
\caption{Confusion matrices on the test set for different models.}
\label{fig:cm}
\end{center}
\end{figure}

\begin{figure*}[t]
    \center{\includegraphics[width=0.9\textwidth]{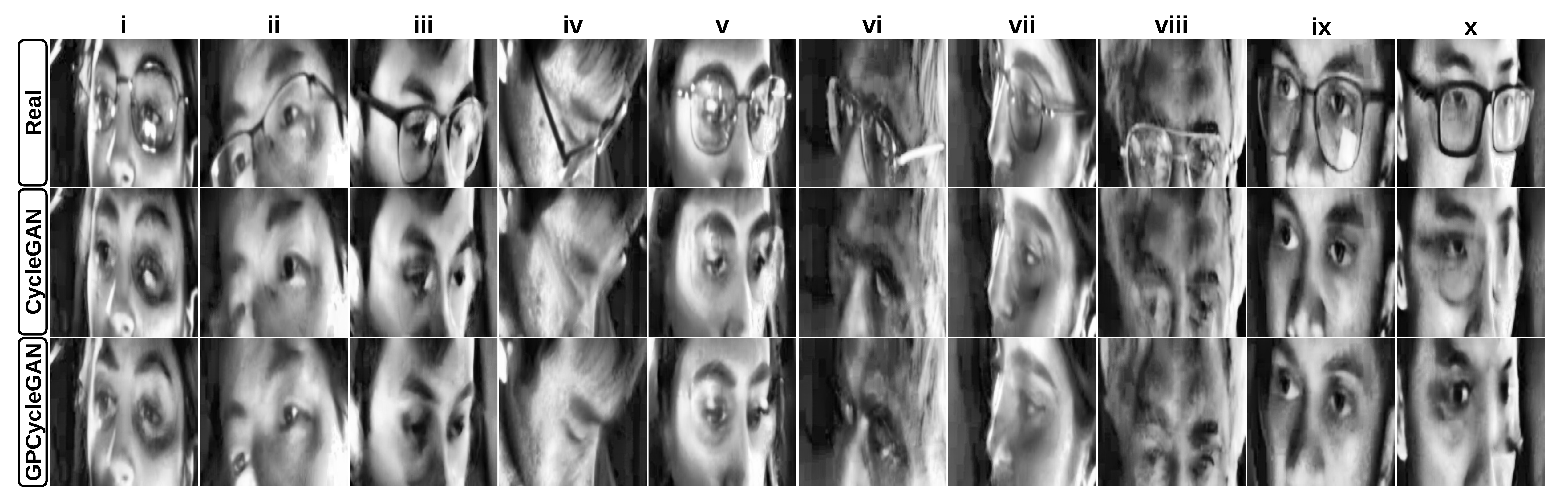}}
    \caption{\label{fig:fake_images} Example real images with eyeglasses, and corresponding generated images after eyeglass removal.}
\end{figure*}

\section{Experiments \& Analyses}
\subsection{Quantitative Metrics \& Results}
Table \ref{tab:test} presents the cross-subject test accuracies for 6 different models. We use two metrics, namely the macro-average and micro-average accuracy, defined as follows:
\begin{equation}
\text{Macro-average accuracy} = \frac{1}{N} \sum_{i=1}^{N} \frac{(\text{True positives})_{i}}{(\text{Total population})_{i}},
\end{equation}
\begin{equation}
\text {Micro-average accuracy} = \frac{\sum_{i=1}^{N}(\text{True positives})_{i}}{\sum_{i=1}^{N}(\text{Total population})_{i}},
\end{equation}
where $N$ is the number of classes. Micro-average accuracy represents the overall percentage of correct predictions, while macro-average accuracy represents the average of all per-class accuracies. 
So, if the data is balanced, which means the number of images for every class is the same, micro-average accuracy will equal macro-average accuracy. In our case, there are more data in higher accuracy classes, so the micro-average accuracies are higher than macro-average accuracies. Thus, we mainly compare the macro-average accuracy since the dataset is not balanced.

First, we consider model \circled{5} trained only on images without eyeglasses to illustrate the domain gap between images with and without eyeglasses. As can be seen, this model performs poorly on a test set containing images outside its training distribution. Next, model \circled{9} represents the scenario where the eyeglasses are not explicitly modeled. Although it is trained on the entire training set, the resulting accuracies indicate a performance penalty, especially when tested on images with eyeglasses.

Next, we see that adding a pre-processing network to remove eyeglasses such as in models \circled{A} and \circled{B} produces slightly worse accuracies than model \circled{9}. This can be attributed to the fact that the gaze classifier has only been trained on half the training set i.e. images without eyeglasses. However, after fine-tuning using the entire training set (Step 3 in Fig.~\ref{fig:training}), the accuracies for both models \circled{A} and \circled{B} increase considerably. In conclusion, our proposed model \circled{B} demonstrates significant improvement over both the baseline model \circled{9} and the vanilla CycleGAN-based model \circled{A} after fine-tuning. The above evidence implies the benefits of our proposed gaze consistency loss, and demonstrates that the generator resulting from the GPCycleGAN model acts effectively as a pre-processing step for the downstream task of gaze estimation. 

In addition to the test set accuracies, we also present the corresponding confusion matrices of all 6 models in Fig.~\ref{fig:cm}. The error modes of our best performing model (Fig.~\ref{fig:cmbft}) can mostly be attributed to confusion between gaze zones close in physical space (for example, \textit{Forward} versus \textit{Speedometer}), occlusion of the eyes by the eyeglass frame and glare, and the inability to distinguish between looking downwards versus closed eyes.


\subsection{Qualitative Results}~\label{sec:qualitative}

\begin{figure}[t]
    \center{\includegraphics[width=0.95\linewidth]{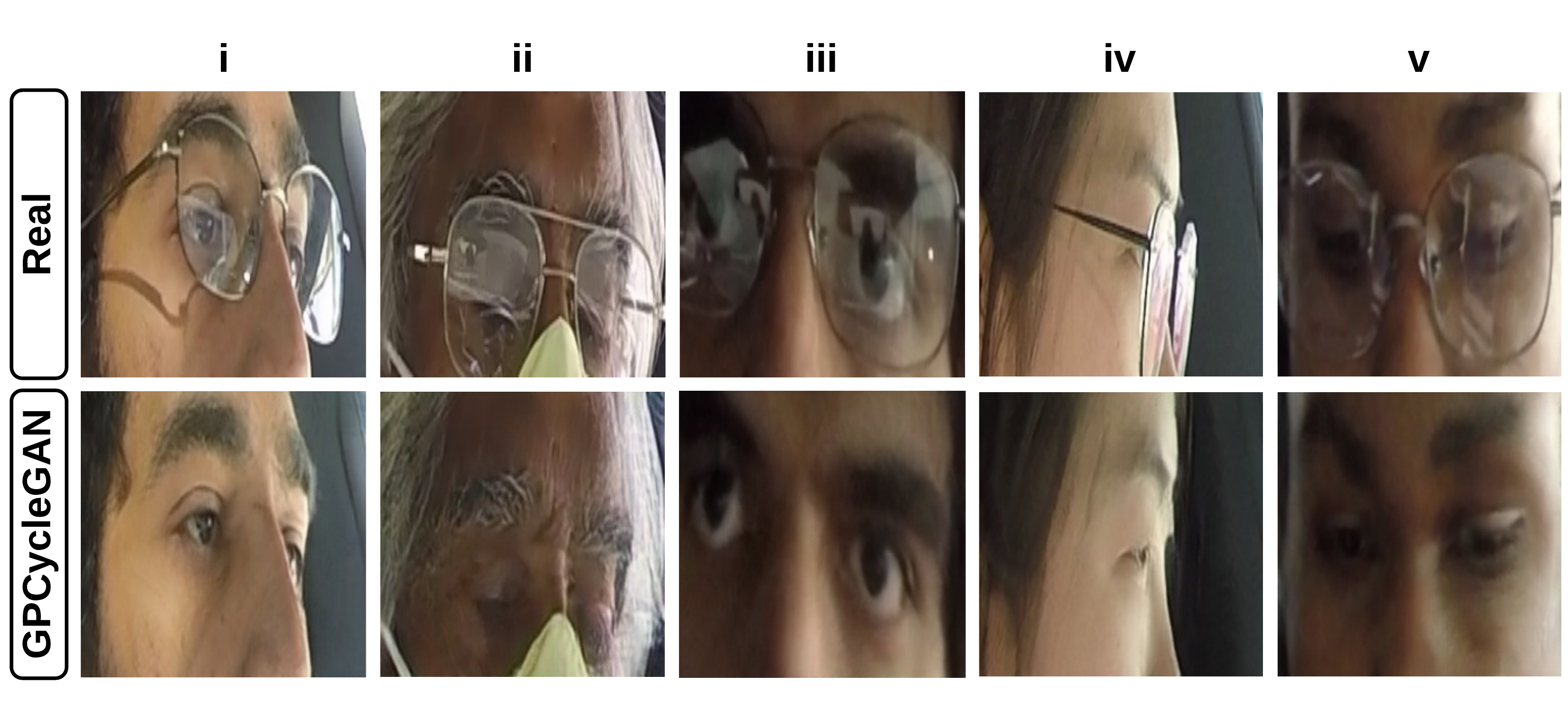}}
    \caption{\label{fig:fake_images_rgb} Example real RGB images with eyeglasses, and corresponding generated images after eyeglass removal.}
\end{figure}

To carry out a qualitative comparison between different GAN variants, we also show 10 examples of eyeglass removal on real images using CycleGAN and GPCycleGAN in Fig.~\ref{fig:fake_images}. In columns i, iii, v, vi, and viii, GPCycleGAN not only removes the eyeglasses, but also removes the glare resulting from it; whereas CycleGAN perceives the glare as part of the sclera. Glare removal is essential for gaze estimation because the glare from glasses is a relatively common occurrence in the real world, and often occludes the eyes, making it harder for models to learn discriminative gaze features. In columns ii, iv, vi, and viii, the images generated from GPCycleGAN are realistic and preserve the gaze more accurately. Columns ii, iii, iv, vi, and vii show that our model does not only work with frontal face images but also performs well for a variety of head poses. Column ix is an example where both models perform well because the gaze is clear and not occluded by the frame or glare. The last column (x) depicts a failure case for both models. The models fail because the frame of the eyeglass is too thick, and both the frame and glare occlude the eye regions severely. These problems could potentially be solved by collecting more data with thicker eyeglass frames, increasing the image resolution, and/or by designing better GANs. 

To test the viability of our approach on different data types, we repeated our prescribed training procedure using an RGB dataset collected for the same task. Fig.~\ref{fig:fake_images_rgb} depicts qualitative results for eyeglass removal using our RGB model. For this RGB dataset, we notice that the resulting images after eyeglass removal tend to be sharper, more realistic, and preserve more details. We believe this is because of the higher resolution and better signal-to-noise ratio of the RGB sensor used to capture the dataset. We provide links to download both IR and RGB datasets in our repository.

\subsection{Comparison with the Other Methods on Different Datasets}

To further demonstrate the generality of our approach, we train and test it on two popular datasets - the Columbia gaze dataset~\cite{ColumbiaDataset2013}, and the MeGlass dataset~\cite{MeGlass2018}.

\begin{figure}[t]
    \center{\includegraphics[width=0.9\linewidth]{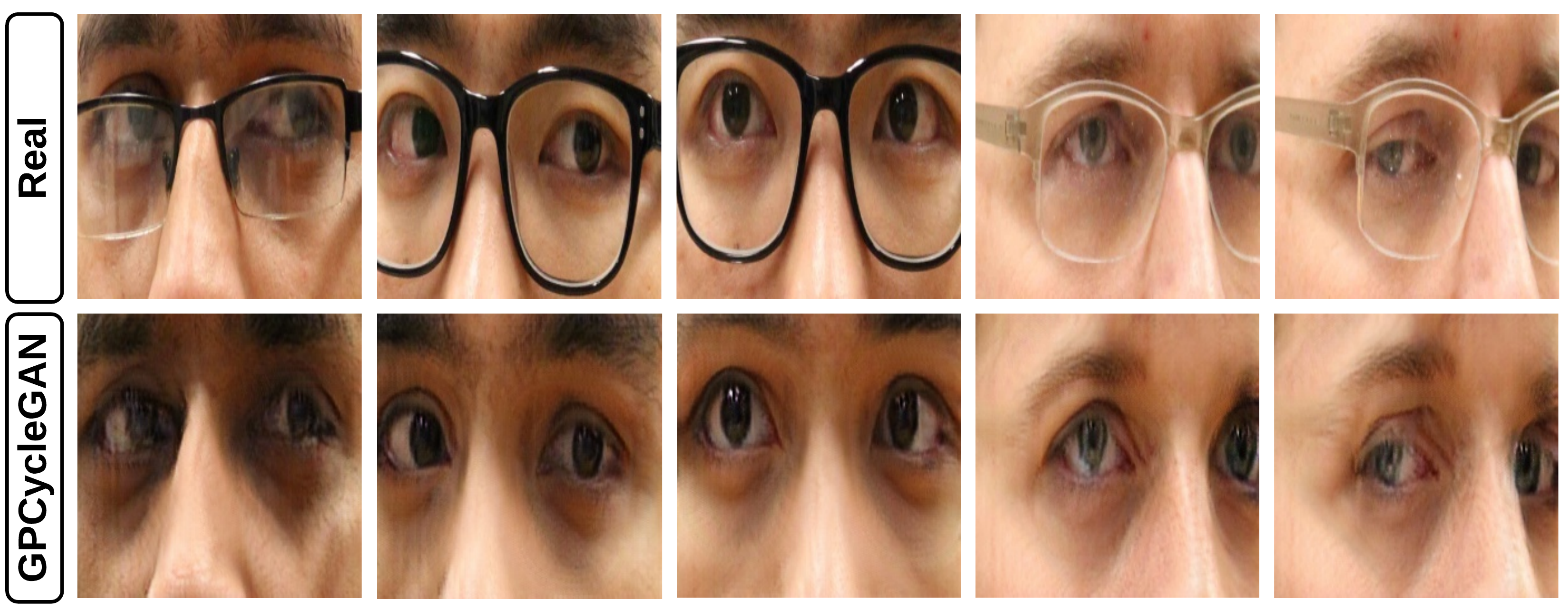}}
    \caption{\label{fig:columbia} Example real RGB images with eyeglasses from the Columbia gaze dataset, and corresponding generated images after eyeglass removal.}
\end{figure}

\textbf{Results on the Columbia Gaze Dataset.} We split the dataset such that the train and test splits contain 45 and 11 subjects respectively. Of the 21 subjects who wore prescription glasses, 6 were included in the test set and the remaining in the train set. The Columbia dataset has very few images to train a gaze classifier (5880 images for 21 gaze classes). So we instead used our pre-trained gaze classifier from the RGB dataset (Section~\ref{sec:qualitative}) to provide the gaze consistency loss for the GPCycleGAN. The results from this model on the test split are shown in the Fig.~\ref{fig:columbia}. Despite using a gaze classifier trained on a different dataset, we can see that the resulting images preserve the gaze direction in the Columbia gaze dataset.

\begin{figure}[t]
    \center{\includegraphics[width=0.95\linewidth]{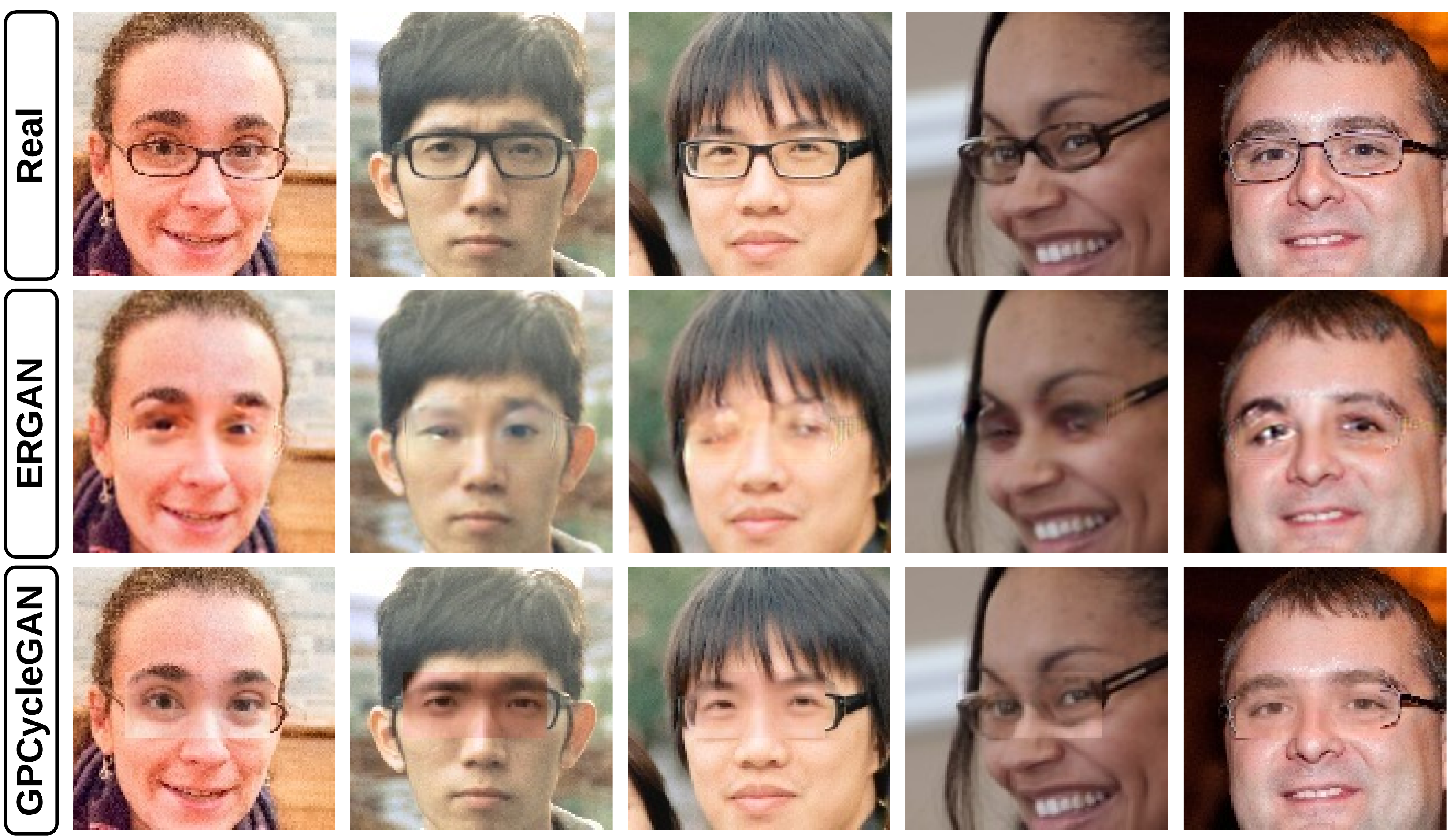}}
    \caption{\label{fig:meglass} Example real RGB images with eyeglasses from the MeGlass dataset, and corresponding generated images after eyeglass removal. We compare the proposed GPCycleGAN approach with ERGAN.}
\end{figure}

\textbf{Results on the MeGlass Dataset.} For this experiment, we split the dataset based on person identities. Of all 1710 identities in the dataset, we assign 1197 to the train split and 513 to the test split. This ensures cross-subject testing in our evaluation. Similar to the Columbia dataset, as the MeGlass dataset does not have labelled gaze directions, we simply use our pre-trained gaze classifier to train the GPCycleGAN model. Since there is no prior work on eyeglass removal for gaze estimation, we compare our approach with the recently proposed eyeglasses removal model ERGAN~\cite{HuERGAN2020} (Fig.~\ref{fig:meglass}). To aid comparison, we overlay our eye crop images after eyeglass removal on the original face image. From this comparison, we can see that ERGAN fails to preserve the gaze direction as it does not enforce any constraints on the gaze. On the other hand, our approach can preserve gaze despite using a gaze classifier trained on a completely different dataset. This also illustrates the benefits of applying eyeglass removal models to eye crop images rather than entire face images, especially if the downstream task is gaze estimation. We believe this is a more efficient use of the generator's modelling capacity.

\subsection{Analyzing the Class Activation Maps}

\begin{figure*}[ht]
    \center{\includegraphics[width=0.95\textwidth]{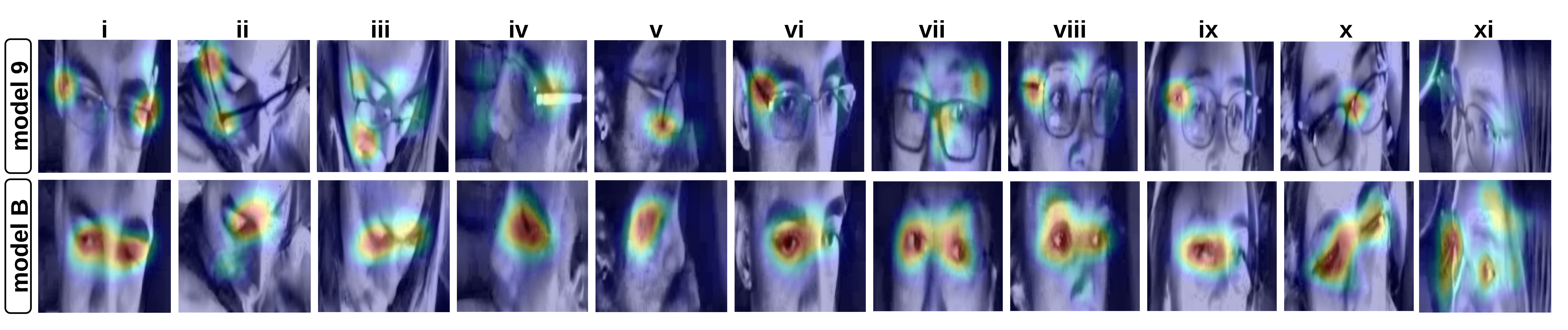}}
    \caption{\label{fig:cams} IR eye image crops from the test set overlaid with ground truth CAMs produced by the gaze classifier. \textbf{Top row:} raw eye image crops with ground truth CAMs produced by model \circled{9}. \textbf{Bottom row:} same eye image crops after eyeglass removal and updated ground truth CAMs produced by model \circled{B}.}
\end{figure*}

Observing the Class Activation Maps (CAMs) produced by classifier models is a simple and effective way of interpreting CNNs and assessing their robustness. In this spirit, we visualize and contrast the CAMs from our proposed approach (model \circled{B}) and the baseline solution (model \circled{9}). This visualization is presented in Fig.~\ref{fig:cams}, where eye image crops from the test set are overlaid with the corresponding ground truth CAMs i.e. the CAMs corresponding to the correct gaze zone. In particular, we compare CAMs generated when raw eye image crops with eyeglasses are directly fed to a gaze classifier (as in model \circled{9}) with CAMs that are obtained by first removing eyeglasses from the eye crop images using a pre-processing network, then feeding this processed image to a fine-tuned gaze classifier (as in model \circled{B}). The CAMs clearly indicate that aside from being superior in terms of quantitative metrics, our proposed model exhibits better interpretability, robustness, and more accurately localizes the eye regions. The baseline model - despite being trained on images with eyeglasses - fails to learn the true nature of the task, and instead overfits to unrelated visual cues. This portends better generalization for our model in comparison to the baseline.

\section{Concluding Remarks}
Reliable and robust gaze estimation on real-world data is essential yet hard to accomplish. A driver's gaze is especially important in the age of partially automated vehicles as a cue for gauging driver state/readiness. In this study, we improved the robustness and generalization of gaze estimation on real-world data captured under extreme conditions, such as data with the presence of eyeglasses, harsh illumination, nighttime driving, significant variations of head poses, etc. For dealing with issues arising from bad lighting, we demonstrate that using an IR camera with suitable equalization/normalization suffices. For images that include eyeglasses, we present eyeglass removal as a pre-processing step using our proposed Gaze Preserving CycleGAN (GPCycleGAN). The GPCycleGAN enables us to train a generator that is capable of removing eyeglasses while retaining the gaze of the original image. This ensures accurate gaze zone classification by a downstream SqueezeNet model. We show that this combined model outperforms both the baseline approach and the vanilla CycleGAN + SqueezeNet model considerably. We also provide example results and illustrations to demonstrate the superiority of our model in terms of interpretability, robustness, and generality. Future work entails improving on the architectures of different components like generator, discriminator, and gaze classifier.


\section{Acknowledgments}
We would like to thank the Toyota Collaborative Safety Research Center (CSRC) for their generous and continued support. We would also like to thank our colleagues for their useful inputs and help in collecting and labeling the dataset. 
Finally, we greatly appreciate all comments and suggestions from the reviewers, associate editor and editor-in-chief, which have improved the overall quality and readability of this work.

\bibliographystyle{IEEEtran}
\bibliography{root}

\begin{IEEEbiography}[{\includegraphics[width=0.96in,height=1.25in,clip,keepaspectratio]{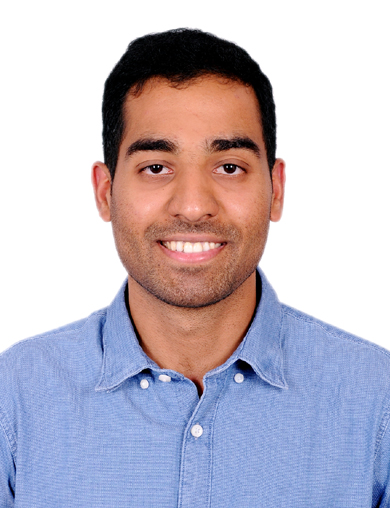}}]{Akshay Rangesh}
is a PhD candidate in electrical engineering from the University of California at San Diego (UCSD), with a focus on intelligent systems, robotics, and control. His research interests span computer vision and machine learning, with a focus on object detection and tracking, human activity recognition, and driver safety systems in general. He is also particularly interested in sensor fusion and multi-modal approaches for real time algorithms.
\end{IEEEbiography}

\begin{IEEEbiography}[{\includegraphics[width=1.1in,height=1.1in,clip,keepaspectratio]{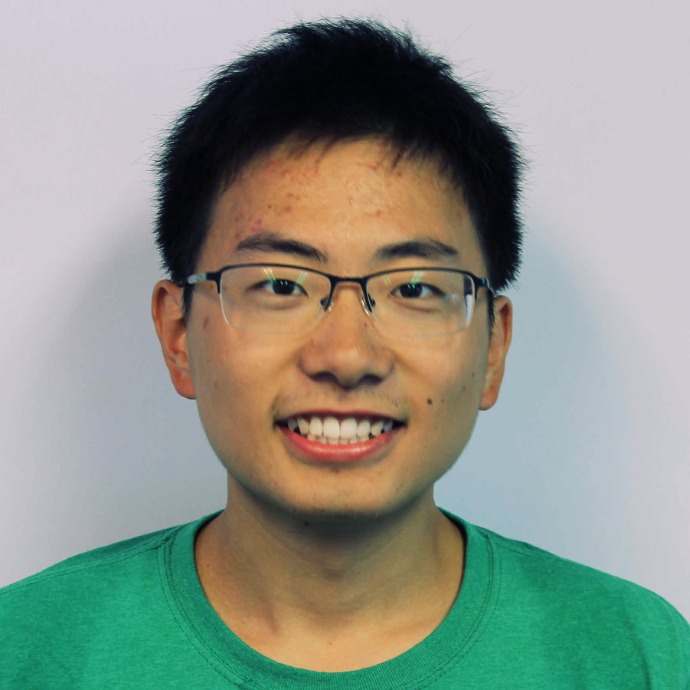}}]{Bowen Zhang}
completed his Masters in Electrical Engineering from the University of California at San Diego (UCSD), with a focus on machine learning and data science. His research interests include human activity recognition, embedded systems and sensing, and human-computer interaction. He is currently a Ph.D. student at the University of California at Santa Barbara (UCSB).
\end{IEEEbiography}

\begin{IEEEbiography}[{\includegraphics[width=1in,height=1.15in,clip,keepaspectratio]{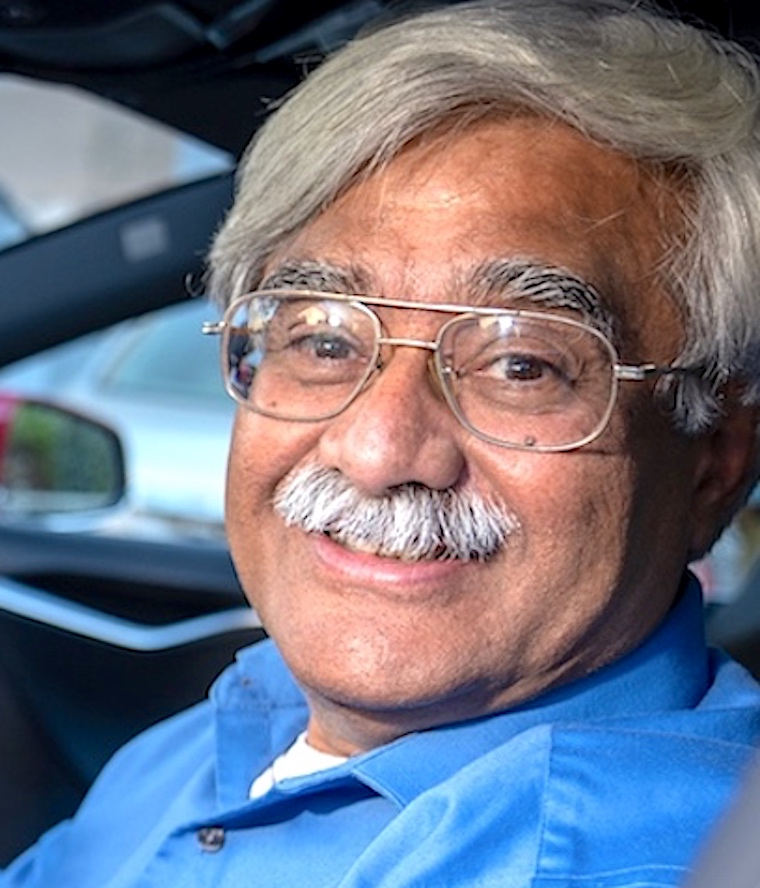}}]{Mohan Manubhai Trivedi}
Mohan M. Trivedi (Life \textit{Fellow} IEEE, SPIE, and IAPR) is a Distinguished Professor of Electrical and Computer Engineering at the University of California, San Diego and the founding director of the Computer Vision and Robotics Research (CVRR, \textit{est. 1986}) and the Laboratory for Intelligent and Safe Automobiles (LISA, \textit{est. 2001}). His research is in intelligent vehicles, intelligent transportation systems (ITS), autonomous driving, driver assistance systems, active safety, human-robot interactivity, machine vision areas. UCSD LISA was awarded the \textit{IEEE ITSS Lead Institution Award}. Trivedi has served as the editor-in-chief of the \textit{Machine Vision and Applications}, Senior Editor of the IEEE Trans on IV and ITSC, as well as Chairman of the Robotics Technical Committee of IEEE Computer Society and Board of Governors of the IEEE ITSS and IEEE SMC societies.
\end{IEEEbiography}

\end{document}